\documentclass[10pt,twocolumn,letterpaper]{article}

\usepackage[pagenumbers]{wacv} 

\usepackage{graphicx}
\usepackage{amsmath}
\usepackage{amssymb}
\usepackage{booktabs}
\usepackage{xcolor}
\usepackage{comment}

\usepackage[pagebackref,breaklinks,colorlinks]{hyperref}

\usepackage[capitalize]{cleveref}
\crefname{section}{Sec.}{Secs.}
\Crefname{section}{Section}{Sections}
\Crefname{table}{Table}{Tables}
\crefname{table}{Tab.}{Tabs.}

\def\wacvPaperID{2577} 
\def\confName{WACV}
\def\confYear{2024}

\begin{document}

\title{GraFT: Gradual Fusion Transformer for Multimodal Re-Identification}

\author{Haoli Yin \and Jiayao (Emily) Li \and Eva Schiller \and Luke McDermott \and Daniel Cummings\\
Modern Intelligence\\
{\tt\small \{haoli, emily.li, eva, luke, daniel\}@modernintelligence.ai}
}
\maketitle

\begin{abstract}
    Object Re-Identification (ReID) is pivotal in computer vision, witnessing an escalating demand for adept multimodal representation learning. Current models, although promising, reveal scalability limitations with increasing modalities as they rely heavily on late fusion, which postpones the integration of specific modality insights. Addressing this, we introduce the \textbf{Gradual Fusion Transformer (GraFT)} for multimodal ReID. At its core, GraFT employs learnable fusion tokens that guide self-attention across encoders, adeptly capturing both modality-specific and object-specific features. Further bolstering its efficacy, we introduce a novel training paradigm combined with an augmented triplet loss, optimizing the ReID feature embedding space. We demonstrate these enhancements through extensive ablation studies and show that GraFT consistently surpasses established multimodal ReID benchmarks. Additionally, aiming for deployment versatility, we've integrated neural network pruning into GraFT, offering a balance between model size and performance.

\end{abstract} 

\section{Introduction}
\label{sec:intro}

Object re-identification (ReID) is the computer vision task of determining whether an object of interest has appeared previously at a distinct place and/or time. At its core, ReID is a matching task, wherein a sampled \textit{query} image is contrasted against a pre-existing \textit{gallery} of images. This task has significant applications in areas such as retail, robotics, multimedia, and surveillance. 
However, ReID comes with significant challenges since the captured representations of objects are subject to a range of uncertainties such as different sensor viewpoints, object poses, occlusions, varying low-resolutions, and environmental conditions \cite{Ye2020ReIDSurvey}. Furthermore, because most conventional ReID algorithms are designed to operate on conventional visible spectrum Red, Green, Blue (RGB) images, there are formidable challenges in less-than-ideal environmental scenarios such as low-light or hazy conditions, similar to the limitations of the human eye. 
To address these issues, additional sensor modalities such as those on the infrared spectrum are commonly used to complement the RGB sensors. However, in the context of multimodal ReID, the critical problem becomes the effective fusion of such diverse data representations, where learning useful object features and the nuances of each modality is essential. 

\begin{figure}[t]
\centering
\includegraphics[scale = .55]{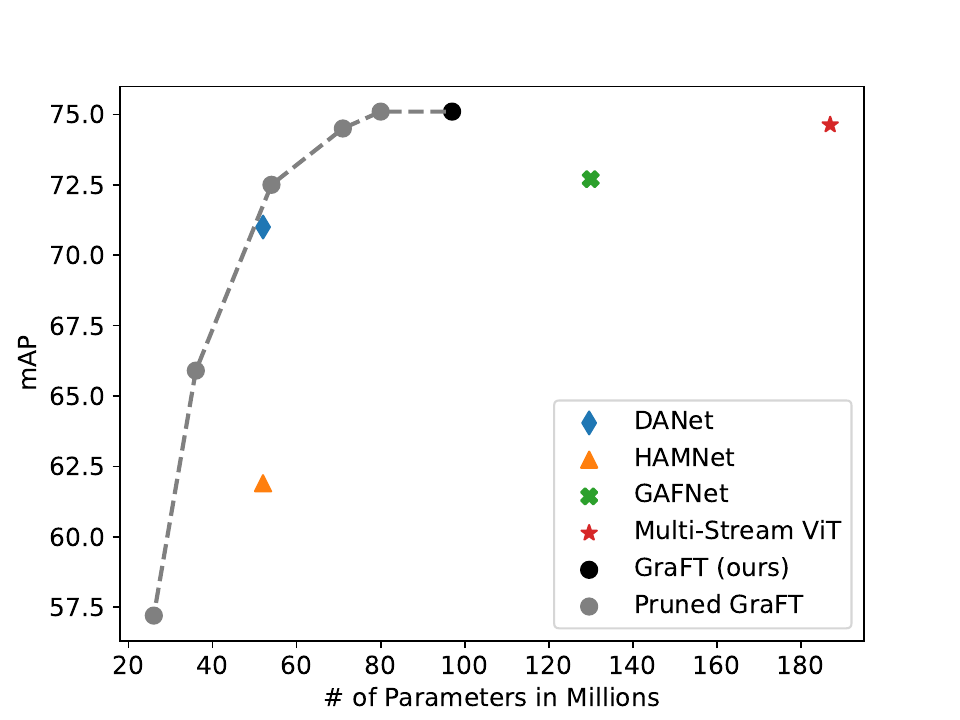}
\caption{Model Size vs Performance on RGBN300 Benchmark}
\label{fig:pareto}

\end{figure}
Deep learning model architectures for multimodal ReID generally fall into two categories: early fusion and late fusion. Early fusion involves the concatenation of images from different modalities, which are then jointly processed through the model. This aims to allow for a more combined understanding of the scene but has the trade-off of sacrificing modality-specific information. Consequently, the richness and depth that each modality offers might not be harnessed to its fullest potential. In contrast, late fusion involves processing each modality individually, and subsequently combining the respective embeddings towards the output of the model. This approach ensures the modality-specific information is independently learned, but may sacrifice object-specific understanding and suffer from impractical model size increases a new modalities are added. 

To address the limitations inherent to both early and late fusion, we propose an effective multimodal learning solution that takes a \textit{gradual} approach to fusion to preserve both modality and object-specific features throughout the model. Our core contributions can be summarized as follows:
\begin{itemize}
    \setlength\itemsep{-0.5ex}
    \item We propose a \textbf{Gradual Fusion Transformer (GraFT)} architecture for multimodal ReID that uses learnable fusion tokens to guide self-attention across encoder layers to extract modality \textit{and} object-specific features.
    \item We develop a unique combination of training paradigms including an augmentation to triplet loss for a more effective ReID feature embedding space. 
    \item We conduct extensive experiments and ablation studies on GraFT using the multimodal ReID benchmark datasets RGBNT100 and RGBN300 \cite{Li-2020a} to outperform existing methods as seen in Fig. \ref{fig:pareto}.
    \item To maximize deployment flexibility, we integrate a neural network pruning capability to allow for a variety of GraFT model size and performance options. 
\end{itemize}

\section{Related Works}

\subsection{Re-Identification}
\begin{figure}
     \centering
     \begin{subfigure}[b]{0.15\textwidth}
         \centering
         \includegraphics[width=\textwidth]{{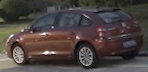}}
         \includegraphics[width=\textwidth]{{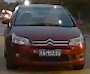}}
         \caption{RGB}
         \label{RGBNT100_RGB}
     \end{subfigure}
     \hfill
     \begin{subfigure}[b]{0.15\textwidth}
         \centering
         \includegraphics[width=\textwidth]{{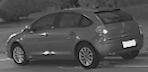}}
         \includegraphics[width=\textwidth]{{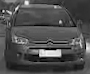}}
         \caption{NIR}
         \label{RGBNT100_NIR}
     \end{subfigure}
     \hfill
     \begin{subfigure}[b]{0.15\textwidth}
         \centering
         \includegraphics[width=\textwidth]{{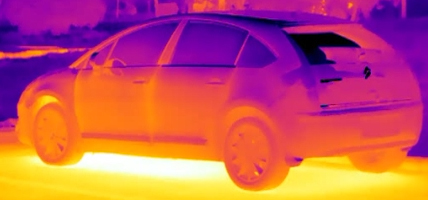}}
         \includegraphics[width=\textwidth]{{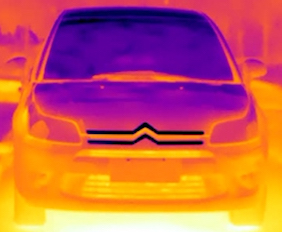}}
         \caption{TIR}
         \label{RGBNT100_TIR}
     \end{subfigure}
        \caption{Example images from the RGBNT100 dataset \cite{Li-2020a}.}
        \label{RGBNT100_figure}
\end{figure}

In recent years, deep learning has rapidly pushed unimodal RGB ReID to new state-of-the-art levels \cite{He_2021_ICCV, Li_Zou_Wang_Xu_Zhao_Zheng_Cheng_Chu_2023, Fu_2021_CVPR, wang2018spatialtemporal, zhu2019viewpointaware, Chen_2023_CVPR, somers2022body, sharma2021person} 
achieving impressive matching accuracy in constrained settings. However, despite this success, relying solely on RGB imagery has inherent weaknesses that present opportunities to explore multimodal techniques. RGB lacks invariance to variations in lighting, occlusion, and viewpoint that commonly occur in uncontrolled real-world ReID scenarios \cite{Zheng_2017_CVPR, peng2022deep, karmakar2021pose}. Furthermore, visibility is significantly degraded under nighttime conditions where illumination is limited \cite{8766119}. In contrast, near-infrared (NIR) and thermal-infrared (TIR) imaging can provide invariant geometric identity cues highly valuable in low light settings \cite{9532715, Li-2020a, tan2023exploring} as shown in Fig. \ref{RGBNT100_figure}. Although a handful of studies have combined RGB and infrared by first individually processing each modality and then concatenating the results together \cite{basaran2019anefficientframework, Li-2020a, ijcai2018p0152, jambigi2021mmd}, 
these approaches do not deeply integrate the complementary modalities architecturally. 
Thus, while unimodal RGB ReID is mature, ample untapped opportunities remain to overcome unimodal limitations by developing principled multimodal fusion approaches. Our proposed GraFT method aims to address these gaps by learning optimized fusion and modality-specific representations to integrate multiple complementary cues for enriched ReID.

\subsection{Multimodal Representation Learning}

Multimodal representation learning aims to integrate information from different data modalities (e.g., images and text) into a joint embedding space. Convolutional neural networks (CNNs) and Vision Transformers (ViTs) are commonly used as visual encoders for this task \cite{mehta2020multimodaldeeplearning}. CNNs apply convolutional filters to hierarchically extract visual features from images while ViTs split images into patches and leverage self-attention, allowing modeling of long-range dependencies.

The key challenge is determining how to effectively combine the unimodal representations from each modality into the joint space. 
Early fusion approaches directly concatenate the raw inputs from each modality before passing them to a joint model. This enables learning cross-modal interactions and aims to create integrated multimodal representations. However, it lacks flexibility since all modalities are handled identically. Late fusion first encodes each modality separately with customized encoders, before concatenating their outputs. This allows architectural optimization tailored to each modality. However, it can miss important joint representations by delaying fusion \cite{barnum2020onthebenefits, zhang2019multimodalintelligence}. Recent methods aim to get the benefits of both approaches through techniques like specialized attention mechanisms or gating to dynamically modulate fusion based on the specific context \cite{shankar2022progressivefusion, Nagrani-2022, gong2023contrastive}. These specialized fusion methods are often designed for particular downstream tasks like visual question answering \cite{faria2023visualquestionanswering} or classification \cite{gong2023contrastive}. However, they do not focus on learning a general joint embedding space that works well for tasks like multimodal person or vehicle re-identification, which is what our proposed solution provides. Our approach focuses on getting a high-quality joint representation for ReID by fusing modalities in a principled manner.

\subsection{Multimodal Re-Identification}

Although the additional of modalities can improve the robustness and performance of ReID, a core challenge remains: how to effectively merge these different data types. Current research suggests that many existing techniques either combine the data too early or too late, compromising the results \cite{jambigi2021mmd, s23094206, kamenou2022closing, guo2022generative, wang2022interact, zheng2022multi, zheng2021robust, Li-2020a}. While much research has been dedicated to merging standard RGB visual data \cite{He_2021_ICCV, Li_Zou_Wang_Xu_Zhao_Zheng_Cheng_Chu_2023, ye2021deep}, there's a clear need for methods that can seamlessly integrate additional visual modalities such as depth and infrared for enhanced identification.

Our approach aims to fill this gap by learning discriminative yet robust embeddings optimized for jointly representing multimodal cues for ReID. We build on top of prior fusion insights and propose innovations to create purpose-built embeddings that efficiently fuse modality-specific information into object-specific features. In particular, we utilize gradual fusion to carefully control the flow of information through transformers and modality encoders to produce robust object representations while maintaining useful specific information from each modality. By tackling representation learning for efficient and holistic fusion, our method represents a significant advance. Our experiments demonstrate state-of-the-art performance on benchmark multimodal ReID datasets, highlighting the benefits of joint embeddings tailored for unified multimodal matching. The powerful yet efficient embeddings produced by our approach could enable deployment of multimodal reID systems in real-world applications.

\section{Method / GraFT Fusion Technique}
\begin{figure}[h]
\centering
\includegraphics[scale = 1.3]{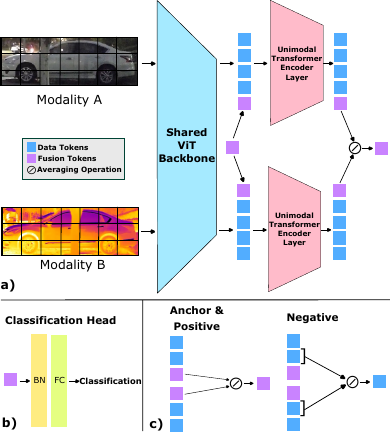}
\caption{GraFT model architecture. a) Main architecture depicting two modality inputs without loss of generality to more modalities. GraFT leverages learnable fusion tokens to gradually fuse together multimodal information. b) The final fusion token embedding is passed to the BNNeck \cite{Luo-2019} and a classifier Fully Connected (FC) layer. c) Our model is trained via a contrastive loss where the fusion tokens are used as the anchor and positive samples and a random data token is used as the negative sample.}
\label{fig:short-b}

\end{figure}

In this section, we describe our proposed method depicted in Fig. \ref{fig:short-b}: GraFT. We first give a high-level overview of our method, then briefly define the popular Vision Transformer (ViT), which serves as a backbone feature extractor. Then, we discuss our token fusion technique, motivate its usage for constructing efficient embedding spaces for Vehicle ReID, and explain in detail the flow of the network from input to output. 

\subsection{Method Overview}
To accomplish gradual fusion, we carefully restrict the flow of information through our model to simultaneously (a) produce a robust object representation and (b) maintain useful and specific information from each modality. First, a transformer backbone extracts features from our raw data, creating information-rich data tokens. We then introduce a learnable fusion token, which is joined with the data tokens and passed separately through each modality's corresponding modality encoder. Through cross-attention, each modality encoder embeds information unique to its respective modality within the fusion token. Finally, the fusion token recombines into one robust object representation through averaging. 

\subsection{Backbone}
The Vision Transformer (ViT) \cite{dosovitskiy2021image} is a derivative of the Transformer architecture \cite{vaswani2023attention}. It adapts the original Transformer's architecture for images by treating them as a sequence of fixed-size patches, equivalent to the tokens in text data. These patches then undergo the same self-attention and feed-forward network operations. In our study, we first use a pretrained ViT-B model on the ImageNet dataset as the feature extractor for ReID tasks, in line with other transformer-based ReID works \cite{He_2021_ICCV}. Empirically, we find that DeiT-B model \cite{DBLP:journals/corr/abs-2012-12877} works the best due to its data distillation pretraining scheme due to a similar data limitation problem, so we adopt that as our final backbone. 

Next, the features extracted from the Vision Transformer (ViT) backbone for each modality are processed. These data tokens are then fed into their corresponding unimodal transformer encoder layers, in conjunction with learnable fusion tokens that are explained in the following section. The final fusion token is then routed to the classification head.


\begin{figure}[t]
\centering
\includegraphics[scale = 2.6]{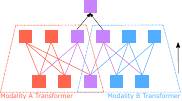}
\caption{Unimodal transformer encoder layer attention flow visualization. Fusion tokens act as the bridge between modalities with modality information being gradually fused through attention.}
\label{fig:attentionflow}

\end{figure}

\subsection{Learnable Fusion Tokens}
To encourage concise communication between features from multiple modalities during fusion while leveraging the self-attention capabilities of Transformers, we employ a learnable fusion token, $\bold{T_f}$ created with Xavier initialization \cite{pmlr-xiavier-init}. We restrict the flow of attention between modalities solely through the fusion token as seen in Fig. \ref{fig:attentionflow}, compelling the token to learn modality-agnostic features at the intersection of all modalities. This method reduces the computational overhead of attention since attention is only required between the fusion tokens and the input sequences from each modality rather than across all modalities.

Additionally, by averaging the transformed outputs for each modality encoder's fusion token, the amount of fusion token parameters required remains constant regardless of the number of modalities, thereby ensuring scalability without compromising fusion proficiency. We also note that similar methods of fusion through learnable tokens have achieved state-of-the-art results on audio-visual discriminative tasks\cite{Nagrani-2022}, but this idea is novel to the ReID task, one that requires the generation of well-ordered embeddings in the vector space. 


\subsection{Architecture Walkthrough}

Given $M$-many modalities, our inputs are tuples of coincidental images for each modality: $(x_1, x_2, ... x_M)$ such that for each image, $x_i \in \mathbb{R}^{C \times H \times W}$ for $i \in [M]$.
Each image is separately fed through a patch embedding that splits the image into $L_d = HW / P^2$ many non-overlapping patches of shape $P^2 C$, such that $(P,P)$ is the resolution of each patch. We use $P=16$ for our base model. To create the patch embeddings, each patch is linearly projected from shape $P^2 C$ to some latent vector size $D$. The shared backbone takes the patched images from each modality and generates useful features as the data tokens, $\bold{T_{d_i}} \in \mathbb{R}^{L_d \times D}$ for some modality $i$. 
\begin{align}
    &\text{Patch Embed}: \mathbb{R}^{C \times H \times W} \rightarrow \mathbb{R}^{L_d \times D}\\
    &\text{Backbone}:  \mathbb{R}^{L_d \times D} \rightarrow \mathbb{R}^{L_d \times D}
\end{align} 

The data and learnable fusion tokens are then concatenated along the sequence length dimension and passed into the modality encoders. The encoders each consist of a multiheaded self-attention module (MHA), and a multilayer perceptron (MLP). For modality $i \in [M]$, we have
\begin{align}
    &\bold{{E_i^1}} = \textit{Concat}(\bold{T_{f}^*}, \bold{T_{d_i}})\\
    &\bold{E_i^2} = \textit{MLP(MHA}(\bold{E_i^1}; \bold{\theta_i}))\\
    &\bold{Z_{f_i}}, \bold{Z_{d_i}} = \textit{Split}(\bold{E_i^{2}})
\end{align} 
such that 
\begin{align*}
\bold{E_i^1}, \bold{E_i^2} \in \mathbb{R}^{(L_d + 1) \times D}, \bold{Z_{f_i}} \in \mathbb{R}^{L_f \times D}, \bold{Z_{d_i}} \in \mathbb{R}^{L_d \times D}
\end{align*}
Note that $\bold{T_{f}^*}$ signifies the expansion or sharing of the fusion token weights at every modality. We then get an aggregate of the fusion tokens from each modality through averaging,
\begin{equation}
    \bold{Z_f} = \frac{1}{M} \sum_{i=1}^M{\bold{Z_{f_i}}}.
\end{equation}
The final embedding for each sample is just the fused token,
\begin{equation}
    \textit{\textbf{Embed}} = \bold{Z_f}
\end{equation}
are subsequently forwarded to the ReID task to calculate distance metrics. During training, $\textit{\textbf{Embed}}$ is also passed to a batch normalization (BN) and fully connected linear layer (FC), both with bias turned off to perform ID classification \cite{Luo-2019}. Applying BN to the embeddings transforms the feature space into a hypersphere centered at the origin with mean zero and unit variance. This standardized distribution aligns with the assumptions of linear models, enabling more effective separation. The resulting spherical input simplifies downstream classification by removing covariate shifts and scaling imbalances between dimensions. However, as only the bias terms of BN and the FC are frozen in our proposed approach, the weights remain trainable via backpropagation. Thus, the model retains some adaptability while benefiting from the normalized feature distribution. 

\section{Training Paradigms}
\subsection{Frozen to unfrozen backbone}

To optimally leverage a backbone pretrained on RGB datasets like ImageNet, we initially employ a frozen pretrained DeIT-B as a general feature extractor. This step encourages the modality-specific encoders and fusion tokens to learn modality-specific features upon general ones and also compensates for the limited training samples available in most ReID datasets. Once the modality encoders are adequately trained, we proceed to fine-tune the general feature extractor to more accurately represent each modality's distribution. We divide training into two stages because using a higher learning rate in the initial stage ensures that the untrained parameters can effectively explore the search space.

\subsection{Contrastive Loss}
Strong clustering within the feature embedding space is an essential characteristic of successful ReID models. To encourage clustering, we employ contrastive loss both between triples and over the entire embedding space.

Between individual triples, which consist of an ``anchor" identity, a ``positive" match, and a ``negative" non-match, we employ soft margin triplet loss \cite{he2021transreid}. Formally, given the triple $\{anchor, positive, negative\}$ with feature embeddings $\{\bold{f_a, f_p, f_n}\}$, soft margin loss $\mathcal{L}_T$ is as follows:
\begin{equation}
\mathcal{L}_T = \log(1 + \exp(\lVert \bold{f_a - f_p}\rVert_2^2) - \lVert \bold{f_a - f_n} \rVert_2^2))
\end{equation}
Over the whole set of feature embeddings, we employ center loss to penalize each embedding's distance from learnable ID-based centroids \cite{wen2016discriminative}. Given the feature embeddings for our anchor identities, $\bold{f_a}$, we compute the cosine distance between each embedding and the learned centroid $\bold{c_y}$ of its corresponding ID $y$. This computation is performed over the minibatch as follows to compute the center loss $\mathcal{L}_C$, where $B$ is batch size. We formalize this as
\begin{equation}
\mathcal{L}_C = \sum_{j=1}^B{\lVert \bold{f_{a_j} - c_{y_j}} \rVert}
\end{equation}
Our complete loss function for ReID combines triplet loss, center loss, and cross-entropy loss with vehicle ID's as classes, $\mathcal{L}_{CE}$, via weighted sum:
\begin{equation}
\mathcal{L}_{total} = \alpha \mathcal{L}_T + \beta \mathcal{L}_C + \gamma \mathcal{L}_{CE}
\end{equation}

For training with a frozen versus a frozen feature extractor, we found best results with ${\alpha = 0.5, \beta = 0, \gamma = 0.5}$ and ${\alpha = 0.5, \beta = 0.0005, \gamma = 0.5}$, respectively. 

\subsection{Augmented Triplet Loss} 

For the anchor and positive ID, we simply use the fused tokens $\mathbf{Z_{f_a}}$ and $\mathbf{Z_{f_p}}$ as the feature embeddings inputted to triplet loss ($\mathbf{f_a}$ and $\mathbf{f_p}$). However, since the fusion tokens are a subset of the data tokens semantically, to encourage effective clustering within classes and optimize for a more linearly separable latent space, we utilize a subset of the data tokens $\mathbf{Z_{d_i}}$ from each modality encoder output $\mathbf{E}_i^2$ as the negative embedding $\mathbf{f_n}$. For each modality, we take a sample token from an arbitrary fixed index (e.g., index 0):
\begin{equation}
    \mathbf{Z}_{d_{i_0}} = Sample(\mathbf{Z}_{d_i})
\end{equation}
    As with the fusion tokens, we then average the data samples across modalities to construct our negative input for triplet soft margin loss, $\mathbf{f_n}$:
\begin{equation}
    \mathbf{f_n} = \frac{1}{M} \sum_{i=1}^M{\mathbf{Z_{d_{i_0}}}}
\end{equation}
We denote augmented triplet loss as Fusion-Fusion-Data (\textit{FFD}), indicating which token is used as the input to anchor positive, and negative, respectively. We similarly define other possible combinations; for example, standard triplet loss, with all samples having fusion token inputs, would be Fusion-Fusion-Fusion (\textit{FFF}). 

\begin{figure}[bt]
\centering
\includegraphics[scale = .62]{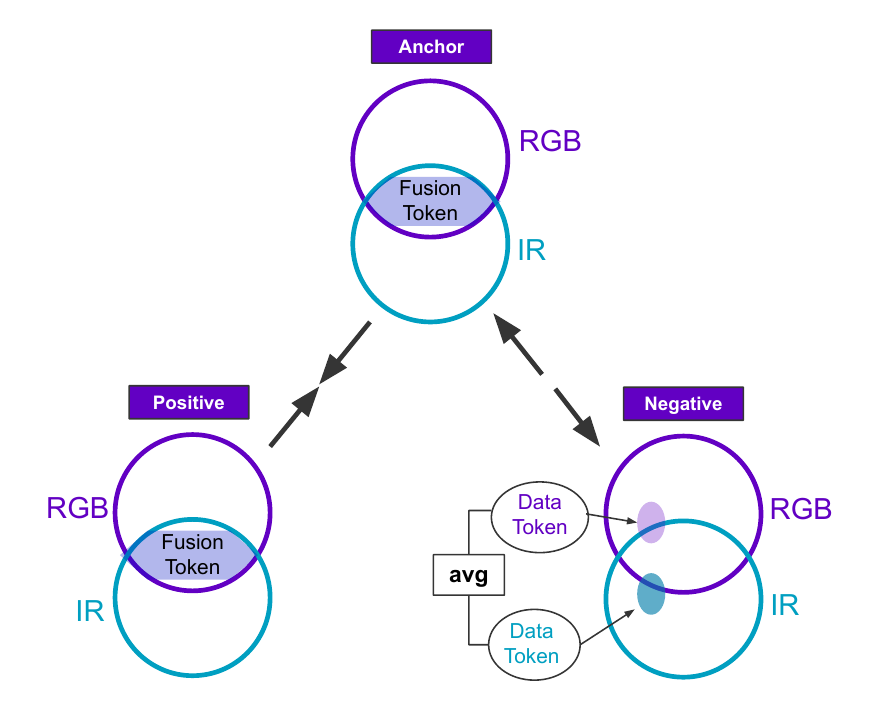}
\caption{Augmented triplet visual explanation using sets. Without loss of generality, RGB and IR represent modality A and B. Fusion tokens act as the intersection of modalities while data tokens are the union of modalities. Through augmented triplet training, anchor and positive samples have their modality intersections pulled together while anchor and negative have the modality intersection pushed away from the modality union.}
\label{fig:augmented_triplet}

\end{figure}

The core benefit of augmented triplet loss lies in the objective of triplet loss: minimizing the distance between the anchor and the positive embedding while maximizing the distance between the anchor and the negative embedding. Directly using the fusion token for the anchor and positive samples in triplet loss awards the design of similar feature embeddings for two samples of the same ID. Meanwhile, using data tokens for the negative ID captures a broader distribution of the negative samples. Essentially, as seen in Fig. \ref{fig:augmented_triplet}, augmented triplet loss leverages the modality intersection to optimize for similar anchor and positive feature embeddings, while leveraging the modality union to further differentiate anchor and negative feature embeddings.

We show empirically in Section 6. that the specific index sampled has no significant impact on model performance, and that reintroducing the data token in the negative ID has a significant positive impact on performance.

\section{Results}

\subsection{Datasets Details}
We train and evaluate our Vehicle ReID method using the benchmark RGBNT100 and RGBN300 dataset \cite{Li-2020a}. The RGBNT100 dataset consists of coincidental RGB, NIR, and TIR coincidental images of 100 unique vehicle IDs from various different camera views. Similarly, the RN300 dataset consists of coincidental RGB and NIR images of 300 unique vehicle IDs from different perspectives.
For RGBNT100, there are a total of 51,750 images with 8,675 in train, 1,715 in query, and 8,575 in \textit{gallery}. For RGBN300, there are a total of 100,250 images with 25,200 in train, 4,985 in \textit{query}, and 24,925 in \textit{gallery}. We sampled triples from the training data randomly and generate multiple sets of triples for every unique image in the dataset to broaden the scope of contrastive loss. To further increase the number of training instances, we sampled multiple positive examples for every unique image sample and set that as a hyperparameter. We found that doing this drastically improved training speed. 

\subsection{Implementation Details}

For software tooling, we use the common deep learning framework PyTorch 2.0.1 \cite{paszke2017automatic}, CUDA 11.6, and Python 3.8. For hardware, we use a cluster of eight A6000 GPUs for distributed training leveraging PyTorch Fabric Lightning \cite{falcon2019pytorch} with the Data Distributed Parallel (DDP) protocol \cite{DBLP:journals/corr/abs-2006-15704}. For the data, we first resize the images to 224x224 pixels then apply horizontal/vertical flips and random erasing \cite{DBLP:journals/corr/abs-1708-04896}. As explained in Section 5.1, we also sample 8 positive examples per anchor to accelerate training. Each mini-batch contains 26 (max amount in GPU virtual memory) anchor, positive, and negative triplets of the respective paired modalities, which are RGB, NIR, and TIR for RGBNT100 and RGB and NIR for RGBN300. For optimization, we utilize the AdamW optimizer \cite{loshchilov2019decoupled} with learning rate and weight decay hyperparameter tuned via optuna \cite{optuna}. For the loss function, we found the best loss hyperparameters to be $\alpha=0.5, \beta=0.005, $ and $\gamma=0.5$. For the architecture, We use only one fusion token to bottleneck the information flow such that the classifier head does not overfit and use one Transformer Encoder Layer provided by the PyTorch library for each unimodal transformer encoder layer.

Inspired by \cite{chen2020cluster-reid}, we leverage two-stage training. For the \textbf{first stage}, we remove center loss as there are no centroids initially to reduce intra-class variation. Through empirical observation and intuition, we find that including the center loss in stage one only slows down training and leads to suboptimal convergence. During stage one training, we also freeze the shared pretrained ViT backbone to only train the unimodel transformer to only train unimodal transformer encoder layers from scratch since those would have large initial losses and take larger steps than the pretrained model would be suited for. To optimizer stage one, we use a learning rate scheduler with linear warmup, square root decay, and finally linear cooldown for the last 10\% of training. For the \textbf{second stage}, we unfreeze the shared pretrained ViT backbone to finetune the entire model and add center loss back in to the overall loss function as there exist class clusters optimized by our augmented triplet loss. For training stage two, we set label smoothing to 0.1 similar to \cite{Luo-2019} and use a constant learning rate at 5e-6.

\subsection{Benchmark Comparisons}

We compare the performance between our gradual fusion methods and other reproducible state-of-the-art (SOTA) results in Tables \ref{table:results_RGBNT100} and \ref{table:results_RGBN300}. These comparison works focus exclusively on the task of multimodal vehicle ReID with specialized architectures tackling each part of this difficult task. These benchmarks all optimize the metrics of R1, R5, and R10, which come from isolated points on a cumulative match curve (CMC). As seen in \cite{zheng2015scalable}, mAP is an aggregate indicator across the CMC curve and indicates whether one algorithm is better at the ReID task as a whole compared to other algorithms. 

\begin{table}
\begin{center}
{\small
\begin{tabular}{l|ccccc}
\toprule
Methods & Params & mAP & R1 & R5 & R10 \\
         &  & (\%) & (\%) & (\%) & (\%) \\
\midrule
HAMNet \cite{Li-2020a}       & 78M & 65.4 & 85.5 & 87.9 & 88.8 \\
DANet \cite{kamenou2022closing}  & 78M & N/A & N/A & N/A & N/A \\
GAFNet \cite{guo2022generative}  & 130M & 74.4 & 93.4 & 94.5 & 95.0 \\
Multi-Stream ViT & 274M & 74.6 & 91.3 & 92.8 & 93.5 \\
\hline
GraFT (Ours) & 101M & \textbf{76.6} & \textbf{94.3} & \textbf{95.3} & \textbf{96.0} \\
\bottomrule
\end{tabular}
}
\end{center}
\caption{ReID performance comparison of GraFT and other methods on the RGBNT100 dataset.}
\label{table:results_RGBNT100}
\end{table}

\begin{table}
\begin{center}
{\small
\begin{tabular}{l|ccccc}
\toprule
Methods & Params & mAP & R1 & R5 & R10 \\
         &  & (\%) & (\%) & (\%) & (\%) \\
\midrule
HAMNet \cite{Li-2020a}       & 52M & 61.9 & 84.0 & 86.0 & 87.0 \\
DANet \cite{kamenou2022closing}        & 52M & 71.0 & 89.9 & 90.9 & 91.5 \\
GAFNet \cite{guo2022generative}      & 130M & 72.7 & 91.9 & 93.6 & 94.2 \\
Multi-Stream ViT & 187M & 73.7 & 91.9 & 94.1 & 94.8 \\
\hline
GraFT (Ours) & 97M & \textbf{75.1} & \textbf{92.1} & \textbf{94.5} & \textbf{95.2} \\

\bottomrule
\end{tabular}
}
\end{center}
\caption{ReID performance comparison of GraFT and other methods on the RGBN300 dataset.}
\label{table:results_RGBN300}
\end{table}

More specifically, we compare against CNN-based works such as HAMNet \cite{Li-2020a}, DANet \cite{kamenou2022closing}, and GAFNet \cite{guo2022generative}. We also compare against a multi-stream ViT baseline, reproduced from \cite{s23094206}, where each modality has its own pretrained ViT-B (as opposed to sharing one) and modality-specific embeddings are fused via averaging to attain the final fused token. 

\subsection{Model Size Analysis}
After the creation of the base GraFT model, we study the performance across different model sizes through the use of neural network pruning \cite{gale2019state}. The aim of neural network pruning is to remove any superfluous parameters, such as individual weights in our scenario, and maintain performance. Multimodal ReID architectures are large vision models that prove to be difficult for inference on smaller devices. Our goal is to create a flexible model for training, yet compressible for inference. We employ magnitude pruning \cite{zhu2017prune} with finetuning at various sparsities to explore how our model performs under model size constraints. More specifically, we performed few-shot iterative magnitude pruning on the backbone architecture: pruning weights for each query, key, and value along with the projection layers and MLPs.
\begin{table}
  \begin{center}
    {\small{
    \begin{tabular}{r|llll}
    \toprule
    Parameters & mAP\(\uparrow\) & R1\(\uparrow\) & R5\(\uparrow\) & R10\(\uparrow\) \\
\midrule
\text{101M} & 76.6 & 94.3 & 95.3 & 96.0 \\
76M & 76.4 & 94.0	& 94.4	& 94.6 \\
70M & 75.1 & 94.1	& 94.9	& 95.5 \\
60M & 73.3 & 92.2 & 93.2	& 94.2 \\
46M & 69.0 & 90.2	& 91.1	& 91.7 \\
\bottomrule
\end{tabular}
}}
\end{center}
\caption{Performance of pruned GraFT variants on RGBNT100. The original dense model has 101 million parameters: one backbone with 3 modality encoders.}
\label{table:prune_RGBNT100}
\end{table}

\begin{table}
  \begin{center}
    {\small{
\begin{tabular}{r|llll}
\toprule Parameters  & mAP\(\uparrow\) & R1\(\uparrow\) & R5\(\uparrow\) & R10\(\uparrow\) \\
\midrule
\text{97M} & 76.6 & 94.3 & 95.3 & 96.0 \\
\text{80M} & 75.1 & 91.0 & 91.8 & 92.4 \\
\text{71M} & 74.7 & 90.7 & 91.53 & 92.1 \\
\text{54M} & 74.5 & 90.1 & 91.1 & 91.8 \\
\text{45M} & 72.48 & 88.67 & 90.0 & 91.0 \\
\text{36M} & 65.9 & 84.4 & 84.4 & 86.1 \\
\text{24M} & 56.2 & 77.7 & 79.2 & 80.4 \\
\bottomrule
\end{tabular}
}}
\end{center}
\caption{Performance of pruned GraFT variants on RGBN300. The original dense model has 97 million parameters: one backbone with 2 modality encoders.}
\label{table:prune_RGBN300}
\end{table}
We prove to be scalable yet compressible, even achieving state-of-the-art performance at a lower parameter count compared to other SOTA as seen in Fig. \ref{fig:pareto}. As shown in Tables \ref{table:prune_RGBNT100} \& \ref{table:prune_RGBN300}, our model collapses after compressing it more than 2.5 times smaller. Through fusion and pruning, we are able to deploy GraFT on more realistic hardware for vehicle ReID in the wild, something not particularly feasible for transformer-based multi-stream approaches \cite{s23094206}.

\section{Ablations}
We conduct a set of ablation studies to assess the influence of our architectural decisions, training strategies, and hyperparameters, as well as to evaluate the scalability of GraFT across different modalities.

In our first ablation study, we compare different fusion techniques on the RGBNT100 dataset as shown in Table \ref{table:ablation_fusion}. In this method, features extracted from each modality are concatenated along the sequence length dimension, complemented by a CLS token. These are subsequently introduced to the Transformer encoder layer. For the subsequent decoder, the aggregate CLS token is utilized. In addition, we evaluate a fusion variant that computes the average of all aggregated feature tokens, analogous to the averaging operation we employ for fusion tokens in Equation 6. All three fusion techniques underwent training under identical conditions and stages. As shown in Table \ref{table:ablation_fusion}, our GraFT fusion approach achieves a notable improvement of 16.1 mAP over the conventional fusion and bests the averaging token fusion method by 14.4 mAP. It is noteworthy that the vanilla fusion averaged token method outperforms the vanilla fusion CLS token technique, indicating that an equally weighted linear combination of the aggregated input yields more useful representations than leveraging a CLS token as an aggregate. We postulate that the sub-optimal performance of vanilla fusion stems from its attempt to fuse all features of all modalities simultaneously, lacking a mechanism to enforce inter-modality coordination. This potentially fails to distill modality-agnostic information crucial for the ReID task into a succinct representation. In contrast, the \textbf{GraFT learnable fusion token provides a dynamic mechanism that allows for adaptive inter-modality interactions, which leads to a more efficient and effective representation of the crucial object-specific information for ReID}.

\begin{table}
  \begin{center}
    \small
    \begin{tabular}{l|rrrr}
    \toprule
    Fusion Method & mAP\(\uparrow\) & R1\(\uparrow\) & R5\(\uparrow\) & R10\(\uparrow\) \\
    \midrule
    Vanilla Fusion: CLS Token & 60.3 & 82.1 & 79.8 & 83.2 \\
    Vanilla Fusion: Averaged Token & 62.0 & 83.9 & 85.9 & 87.4 \\
    GraFT: Fusion Token (Ours) & \textbf{76.4} & \textbf{94.0} & \textbf{94.4} & \textbf{94.6} \\
    \bottomrule
    \end{tabular}
  \end{center}
  \caption{Performance of current vanilla fusion approaches compared with GraFT's Fusion Technique on RGBNT100}
  \label{table:ablation_fusion}
\end{table}

Our second ablation study investigates how the GraFT framework scales with the addition of modalities. We first start by looking at each modality separately in a unimodal GraFT setting and then continue by testing different combinations of modalities beyond just the traditional RGB setting. As anticipated, as more modalities are added, the performance of GraFT improves as shown in Tables \ref{table:ablation_modes_1} and \ref{table:ablation_modes_2} across all performance metrics and the benchmark datasets. This highlights that the \textbf{GraFT approach is able to learn representations across modalities in such a way that allows for object and modality specific data characteristics to be efficiently leveraged}. 

\begin{table}
\begin{center}
{\small
\begin{tabular}{l|cccc}
\toprule
Modality & mAP\(\uparrow\) & R1\(\uparrow\) & R5\(\uparrow\) & R10\(\uparrow\) \\
\midrule
T & 38.6 & 63.5 & 71.0 & 74.7 \\
N & 39.6 & 60.4 & 64.2 & 66.5 \\
R & 49.3 & 70.3 & 75.0 & 77.4 \\
R+N & 54.3 & 76.4 & 79.6 & 81.6 \\
N+T & 62.7 & 86.8 & 89.1 & 90.3 \\
R+T & 67.0 & 89.2 & 91.3 & 92.7 \\
R+N+T & \textbf{76.6} & \textbf{94.3} & \textbf{95.3} & \textbf{96.0} \\
\bottomrule
\end{tabular}
}
\end{center}
\caption{Performance of our method across different modalities on the RGBNT100 dataset. T=Thermal-IR, N=Near-IR, R=RBG.}
\label{table:ablation_modes_1}
\end{table}

\begin{table}
\begin{center}
{\small
\begin{tabular}{l|c|ccc}
\toprule
Modality & mAP\(\uparrow\) & R1\(\uparrow\) & R5\(\uparrow\) & R10\(\uparrow\) \\
\midrule
N & 51.3 & 74.1 & 77.2 & 79.5 \\
R & 60.4 & 81.5 & 83.9 & 86.8 \\
R+N & \textbf{75.1} & \textbf{91.4} & \textbf{92.3} & \textbf{92.9} \\
\bottomrule
\end{tabular}
}
\end{center}
\caption{Performance of our method across different modalities on the RN300 dataset. T=Thermal-IR, N=Near-IR, R=RBG.}
\label{table:ablation_modes_2}
\end{table}

In our third ablation study, we compare various fusion and data token combinations in order of anchor, positive, and negative as related to Fig. \ref{fig:augmented_triplet}. Out of all the combinations, we find that the \textit{FFD} contrastive loss scheme most effectively improves performance by a large margin as shown in Table \ref{table:ablation_triplet}, which follows the intuition as explained in Section 4.3. This demonstrates that our \textbf{augmented triplet loss leads to higher performance than triplet loss with standard inputs}, or with any other combination of data token and fusion token inputs. Taking a closer look, only sampling data tokens (\textit{DDD}) outperforms only using fusion tokens (\textit{FFF}), which can be explained since the \textit{DDD} has a much larger distribution of values to conduct contrastive learning on as compared to the averaged \textit{FFF} fusion tokens. However, it is important to note how incorporating fusion tokens in some form (see \textit{FDD, DFF, DFD}) all outperform the \textit{DDD} approach, which indicates that fusion tokens play an important role in our usage of augmented triplet loss for contrastive learning in multimodal ReID.  

\begin{table}
\begin{center}
{\small
\begin{tabular}{l|cccc}
\toprule
Anchor/+/- & mAP\(\uparrow\) & R1\(\uparrow\) & R5\(\uparrow\) & R10\(\uparrow\) \\
\midrule
FFD & \textbf{76.6} & \textbf{94.3} & \textbf{95.3} & \textbf{96.0} \\
DDD & 59.0 & 85.5 & 89.2 & 90.4 \\
FFF & 57.5 & 81.2 & 84.6 & 86.2 \\
FDD & 62.2 & 80.5 & 87.3 & 88.9 \\
DFF & 63.6 & 86.9 & 89.2 & 90.0 \\
DFD & 60.6 & 85.2 & 88.2 & 90.0 \\
DDF & 57.6 & 83.7 & 86.9 & 88.6 \\
FDF & 38.2 & 60.1 & 63.0 & 65.0 \\
\bottomrule
\end{tabular}
}
\end{center}
\caption{Performance of our method across different Triple Loss Approaches where F=Fusion Token, D=Data Token. For example, \textit{FFD} = anchor (Fusion), positive (Fusion), negative (Data).}
\label{table:ablation_triplet}
\end{table}

As a final observation, we experimented with sampling the input for augmented triplet loss from varying locations within the data tokens. We found that the \textbf{specific data token sampled can be chosen arbitrarily and provide extremely consistent results}: with all other factors remaining identical, selecting the data token from indices $\{5, 10, 15, 20, 25\}$ lead to highly similar mAP outcomes with a standard deviation of $\sigma = 0.0006873$. 

\section{Conclusion}

In this paper, we introduce the Gradual Fusion Transformer (GraFT) for multimodal ReID, a cutting-edge architecture that employs learnable fusion tokens to adeptly capture both modality-specific and object-specific features by guiding self-attention across encoders. Our innovative training paradigm, complemented by an augmented triplet loss, optimizes the ReID feature embedding space, resulting in a more robust model. Through extensive experiments and ablation studies on benchmark datasets RGBNT100 and RGBN300, GraFT not only outperforms existing methods but also offers a new standard in reproducibility, addressing a gap in the current state-of-the-art multimodal ReID methods. To further the utility and adaptability of our approach, we have integrated neural network pruning into GraFT, allowing for a balance between model size and performance, aiming for diverse deployment scenarios. As we advance, our aim is to broaden the applicability of this framework to encompass various application settings and data modalities. 

\textbf{Potential Social Impact}. The topic of ReID brings up a nuanced set of trade-offs in terms of societal impacts. On the one hand, ReID has exciting and unique applications in robotics, public safety, criminal investigations, search and rescue, security authentication, and personalized retail experiences. However, potential negative social aspects of ReID and computer vision in general that need to be considered include bias and discrimination from training data, privacy concerns for public spaces, misidentification in criminal cases, and exploitation of data without consent. Ethical implementation under proper regulation, the respect of privacy rights, and transparent deployment of such systems will be crucial for maximizing the benefits of deep learning ReID models in practice.


\newpage

{\small
\bibliographystyle{ieee_fullname}
\bibliography{egbib}
}

\end{document}